\documentclass{article}
\usepackage{ijcai20}

\usepackage{times}
\usepackage{soul}
\usepackage{url}
\usepackage[utf8]{inputenc}
\usepackage[small]{caption}
\usepackage{graphicx}
\usepackage{amsmath}
\usepackage{amsthm}
\usepackage{amsfonts}
\usepackage{booktabs}
\usepackage{algorithm}
\usepackage{algorithmic}
\usepackage{subfiles}
\usepackage{dsfont}
\usepackage{subfigure}
\usepackage{color}
\title{Scalable Gaussian Process Regression Networks}

\author{
}

\author{
Shibo Li$^1$
\and
Wei Xing$^2$\and
Robert M. Kirby$^{1,2}$\and
Shandian Zhe$^1$
\affiliations
$^1$School of Computing, University of Utah\\
$^2$Scientific Computing and Imaging Institute, University of Utah
\emails
\{shibo,kirby,zhe\}@cs.utah.edu,
wxing@sci.utah.edu
}

\begin{document}
\newcommand{\var}{{\rm var}}
\newcommand{\Tr}{^{\rm T}}
\newcommand{\vtrans}[2]{{#1}^{(#2)}}
\newcommand{\kron}{\otimes}
\newcommand{\schur}[2]{({#1} | {#2})}
\newcommand{\schurdet}[2]{\left| ({#1} | {#2}) \right|}
\newcommand{\had}{\circ}
\newcommand{\diag}{{\rm diag}}
\newcommand{\invdiag}{\diag^{-1}}
\newcommand{\rank}{{\rm rank}}
\newcommand{\nullsp}{{\rm null}}
\newcommand{\tr}{{\rm tr}}
\renewcommand{\vec}{{\rm vec}}
\newcommand{\vech}{{\rm vech}}
\renewcommand{\det}[1]{\left| #1 \right|}
\newcommand{\pdet}[1]{\left| #1 \right|_{+}}
\newcommand{\pinv}[1]{#1^{+}}
\newcommand{\erf}{{\rm erf}}
\newcommand{\hypergeom}[2]{{}_{#1}F_{#2}}

\renewcommand{\a}{{\bf a}}
\renewcommand{\b}{{\bf b}}
\renewcommand{\c}{{\bf c}}
\renewcommand{\d}{{\rm d}}  
\newcommand{\e}{{\bf e}}
\newcommand{\f}{{\bf f}}
\newcommand{\g}{{\bf g}}
\newcommand{\h}{{\bf h}}
\renewcommand{\k}{{\bf k}}
\newcommand{\m}{{\bf m}}
\newcommand{\mb}{{\bf m}}
\newcommand{\n}{{\bf n}}
\renewcommand{\o}{{\bf o}}
\newcommand{\p}{{\bf p}}
\newcommand{\q}{{\bf q}}
\renewcommand{\r}{{\bf r}}
\newcommand{\s}{{\bf s}}
\renewcommand{\t}{{\bf t}}
\renewcommand{\u}{{\bf u}}
\renewcommand{\v}{{\bf v}}
\newcommand{\w}{{\bf w}}
\newcommand{\x}{{\bf x}}
\newcommand{\hx}{{\hat{\x}}}
\newcommand{\hf}{{\hat{\f}}}
\newcommand{\hsf}{{\hat{f}}}

\newcommand{\y}{{\bf y}}
\newcommand{\z}{{\bf z}}
\newcommand{\A}{{\bf A}}
\newcommand{\B}{{\bf B}}
\newcommand{\C}{{\bf C}}
\newcommand{\D}{{\bf D}}
\newcommand{\E}{{\bf E}}
\newcommand{\F}{{\bf F}}
\newcommand{\G}{{\bf G}}
\renewcommand{\H}{{\bf H}}
\newcommand{\I}{{\bf I}}
\newcommand{\J}{{\bf J}}
\newcommand{\K}{{\bf K}}
\newcommand{\hK}{\widehat{\K}}
\renewcommand{\L}{{\bf L}}
\newcommand{\M}{{\bf M}}
\newcommand{\MN}{\mathcal{MN}}
\newcommand{\TN}{\mathcal{TN}}
\newcommand{\N}{\mathcal{N}}  
\newcommand{\Acal}{\mathcal{A}}
\newcommand{\Ocal}{\mathcal{O}}
\newcommand{\Dcal}{\mathcal{D}}
\newcommand{\Xcal}{\mathcal{X}}
\newcommand{\Ycal}{\mathcal{Y}}
\newcommand{\Zcal}{\mathcal{Z}}
\newcommand{\Fcal}{\mathcal{F}}
\newcommand{\Vcal}{\mathcal{V}}
\newcommand{\Lcal}{\mathcal{L}}
\newcommand{\Tcal}{\mathcal{T}}
\newcommand{\Gcal}{\mathcal{G}}
\newcommand{\Hcal}{\mathcal{H}}
\newcommand{\Scal}{\mathcal{S}}

\renewcommand{\O}{{\bf O}}
\renewcommand{\P}{{\bf P}}
\newcommand{\Q}{{\bf Q}}
\newcommand{\R}{{\bf R}}
\renewcommand{\S}{{\bf S}}
\newcommand{\T}{{\bf T}}
\newcommand{\U}{{\bf U}}
\newcommand{\V}{{\bf V}}
\newcommand{\W}{{\bf W}}
\newcommand{\X}{{\bf X}}
\newcommand{\hX}{{\hat{\X}}}
\newcommand{\Y}{{\bf Y}}
\newcommand{\Z}{{\bf Z}}
\newcommand{\bepsi}{{\boldsymbol \epsilon}}
\newcommand{\Mcal}{{\mathcal{M}}}
\newcommand{\Wcal}{{\mathcal{W}}}
\newcommand{\Ucal}{{\mathcal{U}}}

\newcommand{\bfLambda}{\boldsymbol{\Lambda}}

\newcommand{\bsigma}{\boldsymbol{\sigma}}
\newcommand{\balpha}{\boldsymbol{\alpha}}
\newcommand{\bpsi}{\boldsymbol{\psi}}
\newcommand{\bphi}{\boldsymbol{\phi}}
\newcommand{\boldeta}{\boldsymbol{\eta}}
\newcommand{\Beta}{\boldsymbol{\eta}}
\newcommand{\btau}{\boldsymbol{\tau}}
\newcommand{\bvarphi}{\boldsymbol{\varphi}}
\newcommand{\bzeta}{\boldsymbol{\zeta}}

\newcommand{\blambda}{\boldsymbol{\lambda}}
\newcommand{\bLambda}{\mathbf{\Lambda}}
\newcommand{\bOmega}{\mathbf{\Omega}}
\newcommand{\bomega}{\mathbf{\omega}}
\newcommand{\bPi}{\mathbf{\Pi}}

\newcommand{\btheta}{\boldsymbol{\theta}}
\newcommand{\bpi}{\boldsymbol{\pi}}
\newcommand{\bxi}{\boldsymbol{\xi}}
\newcommand{\bSigma}{\boldsymbol{\Sigma}}

\newcommand{\bgamma}{\boldsymbol{\gamma}}
\newcommand{\bGamma}{\mathbf{\Gamma}}

\newcommand{\bmu}{\boldsymbol{\mu}}
\newcommand{\1}{{\bf 1}}
\newcommand{\0}{{\bf 0}}

\newcommand{\bs}{\backslash}
\newcommand{\ben}{\begin{enumerate}}
\newcommand{\een}{\end{enumerate}}

 \newcommand{\notS}{{\backslash S}}
 \newcommand{\nots}{{\backslash s}}
 \newcommand{\noti}{{\backslash i}}
 \newcommand{\notj}{{\backslash j}}
 \newcommand{\nott}{\backslash t}
 \newcommand{\notone}{{\backslash 1}}
 \newcommand{\nottp}{\backslash t+1}

\newcommand{\notk}{{^{\backslash k}}}
\newcommand{\notij}{{^{\backslash i,j}}}
\newcommand{\notg}{{^{\backslash g}}}
\newcommand{\wnoti}{{_{\w}^{\backslash i}}}
\newcommand{\wnotg}{{_{\w}^{\backslash g}}}
\newcommand{\vnotij}{{_{\v}^{\backslash i,j}}}
\newcommand{\vnotg}{{_{\v}^{\backslash g}}}
\newcommand{\half}{\frac{1}{2}}
\newcommand{\msgb}{m_{t \leftarrow t+1}}
\newcommand{\msgf}{m_{t \rightarrow t+1}}
\newcommand{\msgfp}{m_{t-1 \rightarrow t}}

\newcommand{\proj}[1]{{\rm proj}\negmedspace\left[#1\right]}
\newcommand{\argmin}{\operatornamewithlimits{argmin}}
\newcommand{\argmax}{\operatornamewithlimits{argmax}}

\newcommand{\dif}{\mathrm{d}}
\newcommand{\abs}[1]{\lvert#1\rvert}
\newcommand{\norm}[1]{\lVert#1\rVert}

\newcommand{\ie}{{\textit{i.e.,}}\xspace}
\newcommand{\eg}{{\textit{e.g.,}}\xspace}
\newcommand{\etc}{{\textit{etc.}}\xspace}
\newcommand{\EE}{\mathbb{E}}
\newcommand{\dr}[1]{\nabla #1}
\newcommand{\VV}{\mathbb{V}}
\newcommand{\sbr}[1]{\left[#1\right]}
\newcommand{\rbr}[1]{\left(#1\right)}
\newcommand{\cmt}[1]{}

\newcommand{\bi}{{\bf i}}
\newcommand{\bj}{{\bf j}}
\newcommand{\bK}{{\bf K}}

\maketitle

\begin{abstract}	
Gaussian process regression networks (GPRN) are powerful Bayesian models for multi-output regression, but their inference is intractable.  To address this issue, existing methods use a fully factorized structure (or a mixture of such structures) over all the outputs and latent functions for posterior approximation, which, however, can miss the strong posterior dependencies among the latent variables and hurt the inference quality. In addition, the updates of the variational parameters are inefficient and can be prohibitively expensive for a large number of outputs. 
To overcome these limitations, we propose a scalable variational inference algorithm for GPRN, which not only captures the abundant posterior dependencies but also is much more efficient for massive outputs. We tensorize the output space and introduce tensor/matrix-normal variational posteriors to capture the posterior correlations and to reduce the parameters.
We jointly optimize all the parameters  and exploit the inherent Kronecker product structure in the variational model evidence lower bound to accelerate the computation. We demonstrate the advantages of our method in several real-world applications.
 
\end{abstract}

\section{Introduction}

Multi-output regression is an important machine learning problem where the critical challenge is to grasp the complex output correlations to enable accurate predictions. Gaussian process regression networks (GPRN)~\cite{wilson2012gaussian} are promising Bayesian models for multi-output regression, which exploit both the structure properties of neural networks  and the flexibility of nonparametric function learning by Gaussian processes (GP)~\cite{williams2006gaussian}. In GRPNs, the outputs are an adaptive linear projection of a set of latent functions; both the latent functions and projection weights are sampled from independent GPs. In this way,
GPRNs can capture input-dependent, highly nonlinear correlations between the outputs, provide heavy-tail predictive distributions and resist overfitting. 

However, a critical bottleneck of GPRN is the inference intractability. To address this issue, existing methods use a fully factorized posterior approximation over the latent functions and projection weights to conduct mean-field variational inference~\cite{wilson2012gaussian}. In light of the  alternating updates, the posterior covariance matrix for each function and weight can be further parameterized by $N$ (rather than $N^2$) parameters ($N$ is the number of training samples)~\cite{nguyen2013efficient}.  In addition, Nguyen and Bonilla (2013) developed nonparametric variational inference that uses a mixture of diagonal Gaussian distributions as the variational posterior of all the latent variables. 


Despite the success of the aforementioned approaches, they can suffer from applications with very high-dimensional outputs, which are common in real world, e.g., MRI imaging prediction and physical simulations. First, the fully factorized posterior (and the mixture of diagonal Gaussian) can miss the strong posterior dependencies within the weights and latent function values, resulting in suboptimal quality. Second, the alternating mean-field updates and the optimization of non-parametric inference have $\Ocal(NK^2D)$ and $\Ocal(QN^2KD)$ time complexity respectively, where $D$, $K$ and $Q$ are the number of the outputs, latent function and mixture components. When $D$ is very large, say, millions, the computational cost can be prohibitively expensive even for moderated $N$ and/or $K$, say, hundreds.


To overcome these problems, we propose a scalable variational inference algorithm that not only better captures the posterior dependency but also is much more efficient for massive outputs. Specifically, we tensorize the output space so that the projection matrix can be converted into a tensor, based on which we propose a tensor-normal distribution as a joint variational posterior for the projection weights. The tensor-normal posterior not only captures the strong posterior dependencies of the weights, but also requires much less covariance parameters --- orders of magnitude less than the output dimension. Similarly, we incorporate a  matrix-normal variational posterior for all the latent function values to capture their posterior dependency and save the parameters. Finally, we jointly optimize the variational evidence lower bound (EBLO), where we use the Kronecker product properties to decompose the expensive log determinant and matrix inverse to further accelerate the computation. As the result, our algorithm can linearly scale to all $N$, $D$ and $K$, i.e., $\Ocal(NDK)$. 

For evaluation, we first examined our method on three small datasets where the existing GPRN inference approaches are available. Our method  shows not only better predictive performance but also a great speed-up. Then we tested our method in two real-world applications with thousands of outputs and the existing GPRN inference algorithms are not feasible. Compared with several state-of-the-art scalable multi-output regression methods, our method almost always achieves significantly better prediction accuracy. Finally, we applied GPRN in a large-scale physical simulation application for one million output prediction. Our method often improves upon the competing approaches by a large margin. 

 \section{Gaussian Process Regression Networks}
 Let us first introduce the notations and background. Suppose we have a set of $N$ multi-output training examples $\Dcal = \{(\x_1, \y_1), \ldots (\x_N, \y_N)\}$, where each output $\y_n$ ($1\le n \le N$) is $D$ dimensional and $D$ can be much larger than $N$. To model multiple outputs,  Gaussian process regression networks (GPRNs)~\cite{wilson2012gaussian} first introduce a small set of $K$ latent functions, $\{f_1(\cdot), \ldots, f_K(\cdot)\}$. Each latent function $f_k(\cdot)$ is sampled from a GP prior~\cite{williams2006gaussian}, a nonparametric function prior that can flexibly estimate various complex functions from data by incorporating (nonlinear) covariance or kernel functions. 
  Next, GPRN introduces a $D \times K$ projection matrix $\W$, where each element $w_{ij}$($1\le i \le D, 1 \le j \le K$) is also considered as a function of the input, and sampled from an independent GP prior. Given an input $\x$, the outputs are modelled by 
 \begin{align}
 \y(\x)  = \W(\x)[ \f(\x) + \sigma_f \bepsi] + \sigma_y \z  \label{eq:gprn-def}
 \end{align}
 where $\f(\x) = [f_1(\x), \ldots, f_K(\x)]^\top$, and $\bepsi$ and $\z$ are random noises sampled from the standard normal distribution. 
 
  If we view each latent function $f_k(\x)$ as an input neuron, GPRN generates the outputs in the same way as neural networks (NNs).  
 Hence GPRN enjoys the structure properties of NNs. Furthermore,  GPRN accommodates  input-dependent (i.e., non-stationary) correlations of the outputs. Given $\W(\cdot)$, the covariance of arbitrary two outputs $y_i(\x_a)$ and $y_j(\x_b)$ is
\begin{align}
k_{y_i, y_j}(\x_a, \x_b) &= \sum_{k=1}^K w_{ik}(\x_a)\kappa_{{\hsf}_k}(\x_a, \x_b) w_{jk}(\x_b) + \delta_{ab} \sigma_y^2 \notag 
\end{align}
where $\delta_{ab} $ is $1$  if $a=b$ and $0$ otherwise, $\kappa_{\hsf_k}(\x_a, \x_b)=\kappa_{f_k}(\x_a, \x_b) + \delta_{ab}\sigma_f^2$, and $\kappa_{f_k}(\cdot, \cdot)$ is the covariance (kernel) function for the latent function $f_k(\cdot)$. The output covariance are determined by the inputs via the projection weights $w_{ik}(\x_a)$ and $w_{jk}(\x_b)$. Therefore, the model is able to adaptively capture the complex output correlations varying in the input space. This is more flexible than many popular multi-output regression models~\cite{alvarez2012kernels} that only impose stationary correlations  invariant to input locations. 

 Now we look into the joint probability of GPRN on the aforementioned training dataset $\Dcal$. Following the original paper~\cite{wilson2012gaussian}, we assume all the latent functions share the same kernel $\kappa_f(\cdot, \cdot)$ and parameters $\btheta_f$, and all the projection weights the same kernel $\kappa_w(\cdot, \cdot)$ and parameters $\btheta_w$. For a succinct representation, we consider a noisy version of each latent function, $\hat{f}_k(\x) = f_k(\x) + \epsilon_k\sigma_f$ where $\epsilon_k \sim \N(0, 1)$. Since $f_k$ is assigned a GP prior, $\hsf_k$ also has a GP prior and the kernel is $\kappa_{\hsf}(\x_a, \x_b) = \kappa_{f}(\x_a, \x_b) + \delta_{ab}\cdot \sigma_f^2$.  We denote the values of  $\hsf_k$ at the training inputs $\X = [\x_1, \ldots, \x_N]^\top $by $\hf_k = [\hsf_k(\x_1), \ldots, \hsf_k(\x_N)]^\top$ and projection weight $w_{ij}$ by $\w_{ij} = [w_{ij}(\x_1), \ldots, w_{ij}(\x_N)]^\top$. Since $\hsf_k(\cdot)$ is sampled from the GP prior, its finite projection follow a multivariate Gaussian prior distribution, $p(\hf_k|\btheta_f, \sigma_f^2) = \N(\hf_j|\0, \K_\hsf)$ where $\K_\hsf$ is a kernel matrix and each element $[\K_\hsf]_{ij} = \kappa_{\hat{f}}(\x_i, \x_j)$. Similarly, the prior of each  $\w_{ij}$ is $p(\w_{ij}|\btheta_w) = \N(\w_{ij}|\0, \K_w)$ where each $[\K_w]_{ij} = \kappa_w(\x_i, \x_j)$. According to \eqref{eq:gprn-def}, given $\{\w_{ij}\}_{1\le i \le D, 1\le j \le K}$ and $\{\hf_k\}_{1 \le k \le K}$, the observed outputs $\Y = [\y_1, \ldots, \y_N]^\top$ are sampled from $p(\Y|\{\w_{ij}\}, \{\hf_k\}) = \prod_{n=1}^N\N(\y_n|\W_n\h_n, \sigma_y^2\I)$ where $\W_n$ is $D \times K$, each $[\W_n]_{ij}=w_{ij}(\x_n)$, and $\h_n = [\hsf_1(\x_n), \ldots, \hsf_K(\x_n)]^\top$. The joint probability then is
 \begin{align}
 &p(\Y, \{\w_{ij}\}, \{\hf_k\}|\X, \btheta_f, \btheta_w, \sigma_f^2, \sigma_y^2) = \prod_{k=1}^K\N(\hf_k|\0, \K_\hsf)\notag \\
 &\cdot \prod_{i=1}^D\prod_{j=1}^K\N(\w_{ij}|\0, \K_w)\prod_{n=1}^N \N(\y_n|\W_n\h_n, \sigma_y^2\I). \label{eq:gprn-ll}
 \end{align}
 
 The inference of GPRN, namely, calculating the exact posterior distribution of the  latent function values and projection weights, $\{\w_{ij}\}$ and $\{\hf_k\}$, and other parameters, is infeasible due to the intractable normalization constant. While we can use Markov-chain Monte-Carlo sampling, it is known to be slow and hard to diagnose the convergence. For more efficient and tractable inference,  current approaches~\cite{wilson2012gaussian,nguyen2013efficient} introduce approximate posteriors that are fully factorized over the projection matrix and latent functions and then conduct mean-field variational inference~\cite{wainwright2008graphical}. Typically, the approximate posterior takes the following form, 
 \begin{align}
 q(\{\w_{ij}\}, \{\hf_k\})=\prod_{k=1}^{K} q(\hf_k) \prod_{i=1}^D\prod_{j=1}^K q(\w_{ij}) \label{eq:mf-post}
 \end{align}
 where each marginal posterior is an $N$ dimensional Gaussian distribution. The mean-field inference enjoys analytical, alternating updates between each $q(\hf_k)$ and $q(\w_{ij})$. In light of the structure of the updates, the covariance of each posterior can be parameterized by just $N$  rather than $N^2$ parameters~\cite{nguyen2013efficient}. The hyper-parameters $\{\btheta_f, \btheta_w, \sigma_f^2, \sigma_y^2\}$ are then estimated by gradient-based optimization of the variational evidence lower bound (ELBO). In ~\cite{nguyen2013efficient}, a nonparametric variational inference approach is also developed to better capture multi-modality, where the variational posterior is a mixture of diagonal Gaussian distribution, 
 \begin{align}
 q(\u) = \frac{1}{Q} \sum_{j=1}^Q \N(\u|\bmu_j, v_j\I) \label{eq:np-post}
 \end{align}
where $\u$ is a vector that concatenate all $\{\w_{ij}\}$ and $\{\hf_k\}$. The variational parameters $\{\bmu_j, v_j\}$ and the hyper-parameters are jointly optimized by maximizing the variational ELBO.

\section{Scalable Variational Inference}
Despite the success of the existing GPRN inference methods, their performance can be limited by oversimplified posterior structures and they can be computationally too costly for high-dimensional outputs.   First, the fully factorized posterior \eqref{eq:mf-post} essentially assumes the projection weights and latent functions are mutually independent as in their prior, and completely ignores their strong posterior dependency arising from the coupled data likelihoods. Although the Gaussian mixture approximation \eqref{eq:np-post} alleviates this issue, the diagonal covariance of each component can still miss the abundant posterior correlations. 
Second, the alternating mean-field updates for \eqref{eq:mf-post} and the optimization for \eqref{eq:np-post} in the nonparametric inference take the time complexity $\Ocal(NK^2D)$ and $\Ocal(QN^2KD)$ respectively in each iteration. When $D$ is very large, say, millions, the computation can be prohibitively expensive even for moderated $N$ or $K$, e.g, hundreds. 


To improve both the inference quality and computational efficiency to massive outputs,  we develop a scalable variational inference algorithm for GPRN, presented as follows. 
\subsection{Matrix and Tensor Normal Posteriors}
First, to capture the posterior dependency between the latent functions,  we use a matrix normal distribution as the joint variational posterior for the $N \times K$ function values $\F = [\hf_1, \ldots, \hf_K]^\top$,
\begin{align}
&q(\F) = \MN(\F, \M, \bSigma, \bOmega ) \notag \\
&= \N(\vec(\F)|\vec(\M), \bSigma \otimes \bOmega ) \label{eq:post-f}
\end{align}
where $\otimes$ is the Kronecker product, $\bSigma$ and $\bOmega$ are $N \times N$ row covariance and $K \times K$ column covariance matrices, respectively. To ensure the positive definiteness, we further parameterize $\bSigma$ and $\bOmega$ by their Cholesky decomposition, $\bSigma = \U\U^\top$ and  $\bOmega = \V\V^\top$. The matrix Gaussian posterior not only captures the dependency of the function values, but also reduces the number of parameters and computational cost. If we use a factorized posterior over each $\hf_k$, the number of parameters is $NK + KN(N+1)/2$, including the mean and Cholesky decomposition of the covariance for each $k$, while our approximate posterior only needs $NK + N(N+1)/2 + K(K+1)/2$ parameters.

Next, we consider the variational posterior of the projection weights $\{\w_{ij}\}$. To obtain a joint posterior yet with compact parameterization, we tensorize the $D$ dimensional output space into an $M$-mode tensor space, $d_1 \times \ldots \times d_M$ where $D= \prod_{m=1}^M d_m$. For simplicity, we set $d_1 = \ldots = d_M =d =  \sqrt[M]{D}$. Then we can organize all the weights into an $N \times K \times d_1 \times \ldots \times d_M$ tensor $\Wcal$. To capture the posterior dependencies between all the weights, we introduce a tensor normal distribution --- a straightforward extension of the matrix normal distribution --- as the variational posterior for $\Wcal$, 
\begin{align}
&q(\Wcal) = \TN(\Wcal|\Ucal, \bGamma_1, \ldots, \bGamma_{M+2}) \notag \\
&=\N(\vec(\Wcal)|\vec(\Ucal), \bGamma_1 \otimes \ldots \otimes \bGamma_{M+2}) \label{eq:post-w}
\end{align}
where $\Ucal$ is the mean tensor, $\{\bGamma_1, \ldots, \bGamma_{M+2}\}$ are covariance matrices in each mode, 
 $\bGamma_1$ is $N \times N$, $\bGamma_2$ is $K \times K$, and $\bGamma_{3:M+2}$ are $d \times d$. \cmt{We parameterize  $\Ucal = \u_1 \circ \ldots \circ \u_{M+2}$ where $\circ$ is the outer-product [?], $\u_1$, $\u_2$, and $\u_{3:M+2}$ are $N$, $K$ and $d$ dimensional vectors respectively. }To ensure positive definiteness, we parameterize each covariance matrix by its Cholesky decomposition, $\bGamma_m = \L_m\L_m^\top$. Note that to represent the entire $NKD \times NKD$ covariance matrix  of $q(\Wcal)$, we only need $N(N+1)/2 + K(K+1)/2 + Md(d+1)/2$ parameters which can be even far less than $NDK$. Take $D = 10^6, N = 100, K = 10$ as an example, the number of the covariance parameters is $0.2\% NKD$ (for $M = 3$). For the mean-field posterior in \eqref{eq:mf-post}, the number of covariance parameters for all the projection weights is $NKD$ even with the compact representation. Therefore, our tensor normal posterior not only preserves the strong correlations among all the weights, but also saves much more parameters and so computation cost. 
 \cmt{
 The number of parameters in our tensor Gaussian posterior is $NKD + N(N+1) /2 + K(K+1)/2 + Md(d+1)/2$. Note that $d = \sqrt[M]{D}$, the number of parameters are orders of magnitude smaller than the output dimension. By contrast, if we use factorized posterior over each $\w_{ij}$, the number of parameters is many times of the output dimension, $NKD + KDN(N+1)/2$, which can be prohibitively costly to estimate for large $D$. Take $D = 10^6, N = 100, K = 10$ as an example, the factorized posterior needs to estimate $\sim10^{11}$ parameters while our tensor Gaussian posterior (for $M=3$) only needs $\sim10^{4}$.  Therefore, our tensor Gaussian posterior not only preserves the strong correlations among all the weights, but also is much more competent for massive outputs regression.  
}

  Finally, we choose our variational posterior  as $q(\F, \Wcal) = q(\F)q(\Wcal)$. While it stills factorizes over $\F$ and $\Wcal$, the strong correlations of many variables within $\F$ and $\Wcal$ are captured, and hence still improves upon the fully factorized posterior.  
\subsection{Simplified Variational Evidence Lower Bound}
 Now, we derive the evidence lower bound (ELBO) with our proposed variational posterior, 
\[
 \Lcal = \EE_{q}[\log\frac{p(\Y, \F, \Wcal|\btheta_f, \btheta_w, \sigma_f^2, \sigma_y^2)}{q(\F, \Wcal)}]  
 \]
 We will  maximize the ELBO to jointly optimize the variational parameters and hyper-parameters. To accelerate the computation, we further use the properties of the Kronecker product~\cite{stegle2011efficient} in our variational posteriors (\eqref{eq:post-f} and \eqref{eq:post-w}) to dispose of their full covariances and derive a much simplified bound, 
 \begin{align}
 \Lcal &= -\mathrm{KL}\big(q(\Wcal) \| p(\Wcal)\big) - \mathrm{KL}\big(q(\F)\|p(\F)\big) \notag \\
 &+ \EE_{q}[ \log p(\Y|\X, \Wcal, \F)] \label{eq:elbo}
 \end{align}
 where
 \begin{align}
 &\mathrm{KL}\big(q(\Wcal) \| p(\Wcal)\big) = \frac{1}{2}\big[\tr(\K_w^{-1}\bGamma_1)\prod_{m=2}^{M+2}\tr(\bGamma_m) + DK\log|\K_w| \notag \\
 &+ \tr(\K_{\hsf}^{-1}\U_1\U_1^\top)   - \sum_{m=1}^{M+2}\frac{NDK}{t_m} \log|\bGamma_m|\big], \notag \\
 &\mathrm{KL}\big(q(\F)\|p(\F)\big) = \frac{1}{2}\big[ \tr(\K_{\hsf}^{-1}\bSigma)\tr(\bOmega) + \tr(\K_{\hsf}^{-1}\M\M^\top) \notag \\
 & + K\log|\K_{\hsf}| - (K\log|\bSigma| + N\log|\bOmega|)  \big], \notag 
 \end{align}
 and
 \begin{align}
 &\EE_{q}[ \log p(\Y|\X, \Wcal, \F)] = -{ND}\log \sigma_y - \sum_{n=1}^N \frac{1}{2\sigma_y^2}\big[\y_n^\top\y_n \notag \\
 & - 2\y_n^\top \EE_{q}[\W_n]\EE_{q}[\h_n]\big] + \tr(\EE_{q}[\W_n^\top\W_n]\EE_{q}[\h_n\h_n^\top])\big]. \notag 
 \end{align}
 Here $\U_1$ is an $N \times DK$ matrix, obtained by unfolding the mean tensor $\Ucal$ at mode $1$, $t_m$ is the dimension of mode $m$ of the tensor $\Wcal$, $\EE_{q}[\W_n]$ is obtained by taking the $n$-th slice of $\Ucal$ at mode $1$ and then reorganize it into a $D \times K$ matrix, $\EE_{q}[\h_n]$ the $n$-th row vector of $\M$, and the remaining moments are calculated by
 \begin{align}
& \EE_{q}[\W_n^\top\W_n] = \bGamma_2 \prod_{m=3}^{M+2}\tr(\bGamma_m) + \EE_{q}[\W_n]^\top\EE_{q}[\W_n],\notag \\
& \EE_{q}[\h_n\h_n^\top] = \bOmega\cdot \mathrm{diag}(\bSigma) + \EE_{q}[\h_n]\EE_{q}[\h_n]^\top. \notag 
 \end{align}
 As we can see, the computation of the ELBO \eqref{eq:elbo} only involves  the covariance matrices at each mode of $\Wcal$, $\F$ and the kernel matrices: $\{\bGamma_{1:M+2}, \bSigma, \bOmega, \K_{\hsf}, \K_w\}$, which are small even for a very large number of outputs. Hence the computation is much simplified. In addition, due to the compact parameterization, optimizing the ELBO is much easier.  We then use gradient-based optimization methods to jointly estimate the parameters of the variational posteriors and the hyper-parameters $\{\sigma_f^2, \sigma_y^2, \btheta_f, \btheta_w\}$. 
 
 \subsection{Prediction}
Given a new input $\x^*$, we aim to use the estimated variational posterior  to predict the output $\y^*$.  The posterior mean of $\y^*$ is computed by $\EE[\y^*|\x^*, \Dcal] = \EE[\W(\x^*)]\EE[\hf(\x^*)] $ where $\hf(\x^*) = [\hsf_1(\x^*), \ldots, \hsf_K(\x^*)]^\top$. Each $[\EE[\W(\x^*)]]_{ij}$ is computed by $\k_w^*\K_w^{-1}\v_{ij}$ where $\k_w^* = [\kappa_w(\x^*, \x_1), \ldots, \kappa_w(\x^*, \x_N)]$ and $\v_{ij}$ is obtained by first reorganizing $\Ucal$ (the posterior mean of $\Wcal$) to an $N \times K \times D$ tensor $\widehat{\Ucal}$ and then taking the fiber $\widehat{\Ucal}(:, i, j)$. Each $\EE(\hsf_k(\x^*)) = \k_\hsf^* \K_{\hsf}^{-1}\M(:, k)$ where $\k_\hsf^* = [\kappa_\hsf(\x^*, \x_1), \ldots, \kappa_\hsf(\x^*, \x_N)]$. 

The predictive distribution, however, does not have a close form, because the likelihood is non-Gaussian w.r.t the projection weights and latent functions. To address this issue, we can use Monte-Carlo approximations. We can generate a set of i.i.d posterior samples of $\W(\x^*)$ and $\hf(\x^*)$, denoted by $\{\widetilde{\W}_t, \widetilde{\f}_t\}_{t=1}^T$, and then approximate $p(y^*|\x^*, \Dcal) \approx \frac{1}{T}\sum_{t=1}^T \N(\y^*|\widetilde{\W}_t \widetilde{\f}_t, \sigma_y^2\I)$. Note that when the output dimension is large, the posterior sample of $\W(\x^*)$ can be too costly to generate. We can instead approximate the predictive distribution of each single output $y^*_j$ with the same method. 
 
 \subsection{Algorithm Complexity}
The overall time complexity of our inference algorithm is $\Ocal(N^3+K^3 + Md^2 + NDK)$. When $D\gg \{N, K\}$ and $Md^2 \le D$, the complexity is $\Ocal(NDK)$. It is trivial to show that when $3\le M \le \sqrt[3]{D}$, we always have  $Md^2 \le D$. \cmt{Actually, in most cases, $MD^{\frac{2}{M}} \ll D$. For example, when $D = 10^6$ and $M=3$, $MD^{\frac{2}{M}} = 0.03D$.} The space complexity is $\Ocal(NKD + ND + N^2 + K^2 + Md^2)$ including the storage of training data, kernel matrices,  the variational posterior parameters and the other parameters. 


\section{Related Work}
Many multi-output regression approaches have been proposed and most of them are based on GPs; see an excellent review in ~\cite{alvarez2012kernels}. A classical method is the linear model of coregionalization (LMC)~\cite{goulard1992linear}, which projects a set of latent functions to the multi-output space; each latent function is sampled from an independent GP.  PCA-GP is a popular LCM~\cite{higdon2008computer} that identifies the projection matrix by Singular Value Decomposition (SVD). The variants of PCA-GP include  KPCA-GP~\cite{xing2016manifold}, IsoMap-GP~\cite{xing2015reduced}, etc. GPRN~\cite{wilson2012gaussian} is another instance of LMC. However, since the projection weights  are also sampled from GPs, the prior of the outputs is no longer a GP. Other approaches include convolved GPs~\cite{higdon2002space,boyle2005dependent,alvarez2019non} and multi-task GPs~\cite{bonilla2007kernel,bonilla2008multi,rakitsch2013all}. Convolved GPs generate each output by convolving a smoothing kernel and a set of latent functions. Multi-task GPs define a product kernel over the input features and task dependent features (or free-from task correlation matrices). Despite their success, both types of models might be too costly ($\Ocal((ND)^3)$ or $\Ocal(N^3 + D^3)$ time complexity) for high-dimensional outputs. To mitigate this issue, several sparse approximations have been developed~\cite{alvarez2009sparse,alvarez2010efficient}. Recently, Zhe \textit{et al.} (2019) introduced latent coordinate features in the tensorized output space to model complex correlations and to predict tensorized outputs. Their method essentially constructs a product kernel with which to simplify the inference.  By contrast, our work introduces tensor/matrix-normal variational posteriors to improve GPRN inference and does not assume any special kernel structure in GPRN. 

\section{Experiment}
\begin{figure*}[htbp]
	\centering
	\subfigure[\textit{Jura}]{
		\begin{minipage}[t]{0.33\linewidth}
			\centering
			\includegraphics[width=2in]{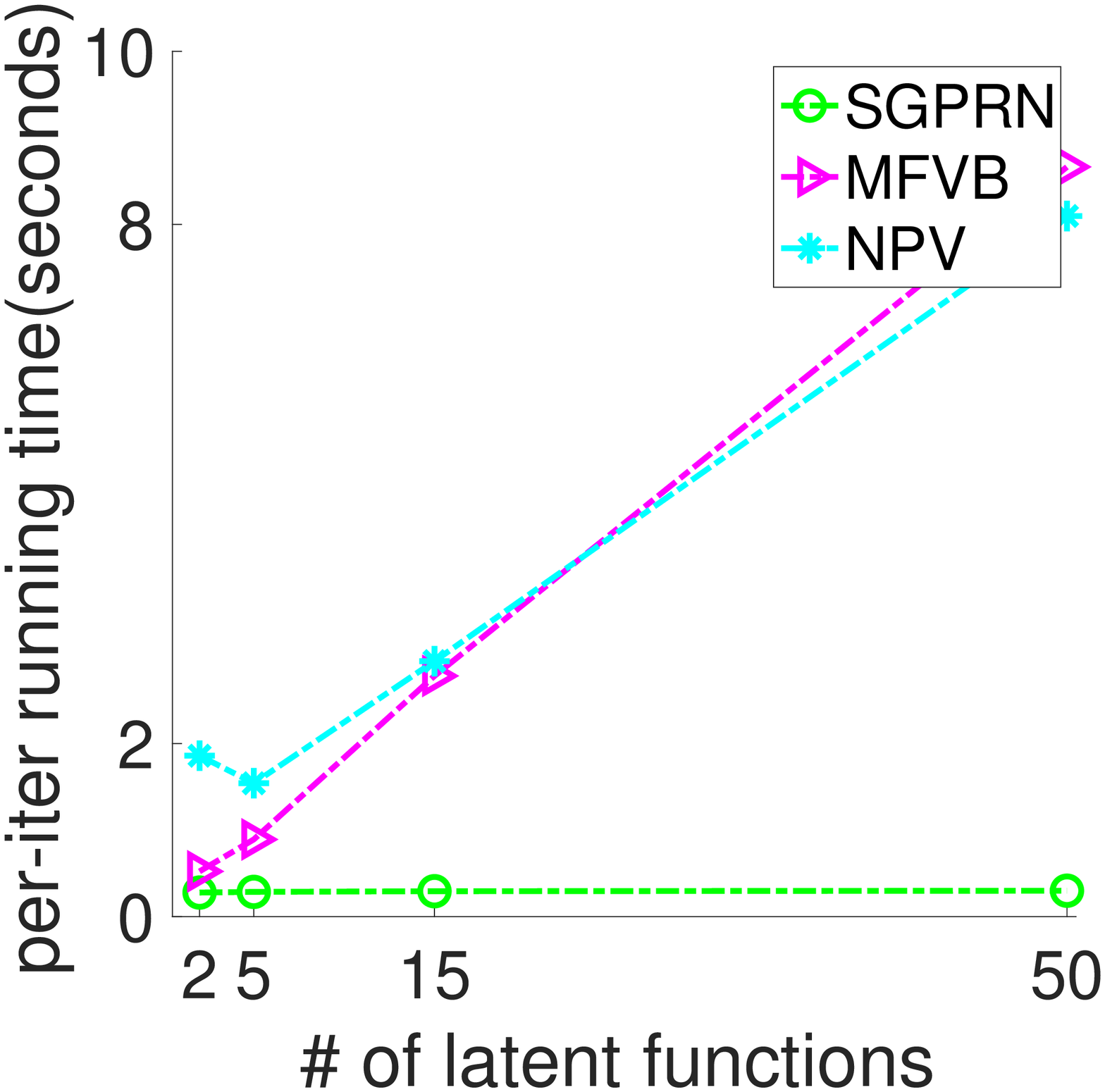}
		\end{minipage}%
	}%
	\subfigure[ \textit{Equity}]{
		\begin{minipage}[t]{0.33\linewidth}
			\centering
			\includegraphics[width=2in]{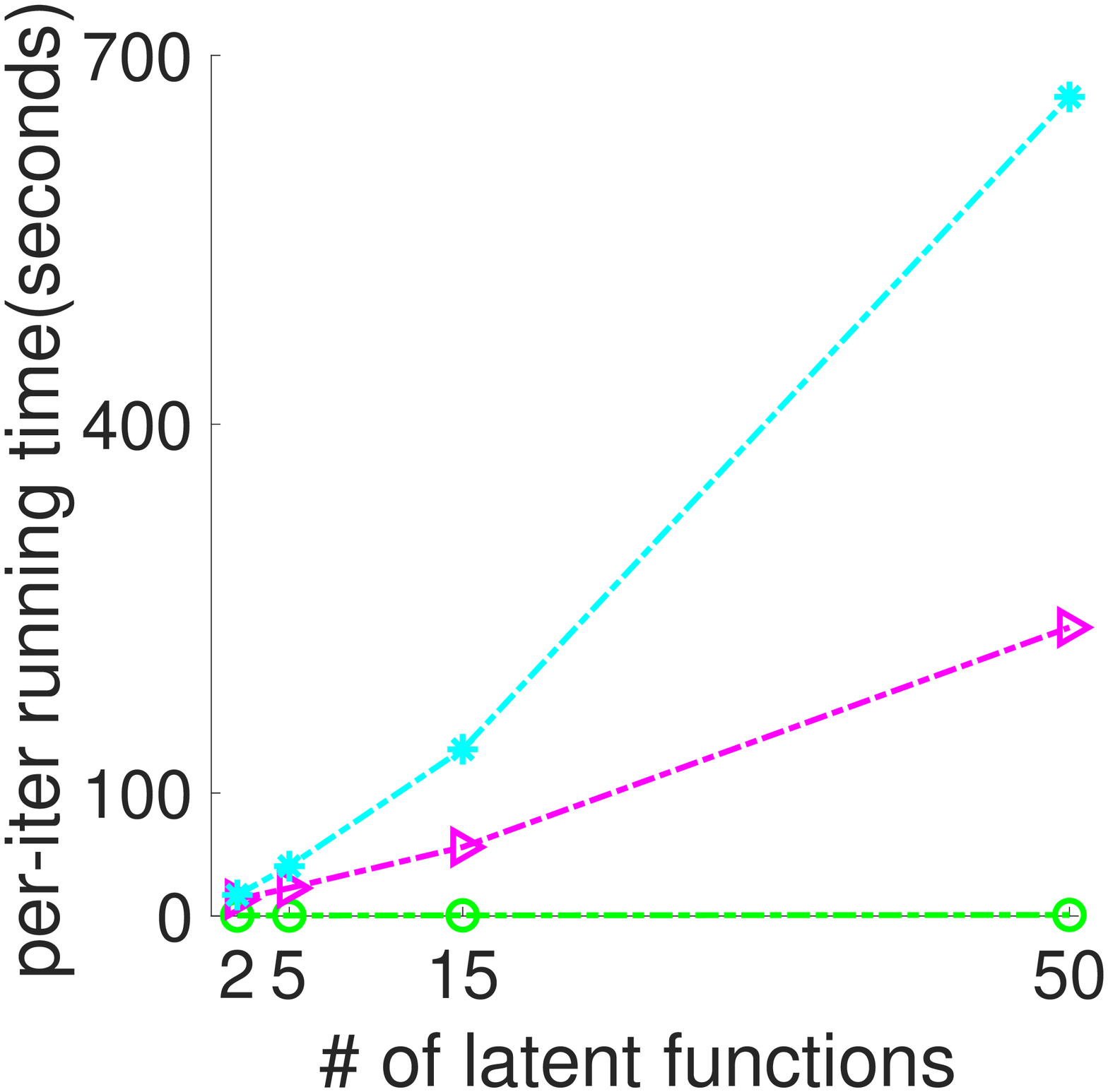}
		\end{minipage}%
	}%
	\subfigure[\textit{PM 2.5}]{
		\begin{minipage}[t]{0.33\linewidth}
			\centering
			\includegraphics[width=2in]{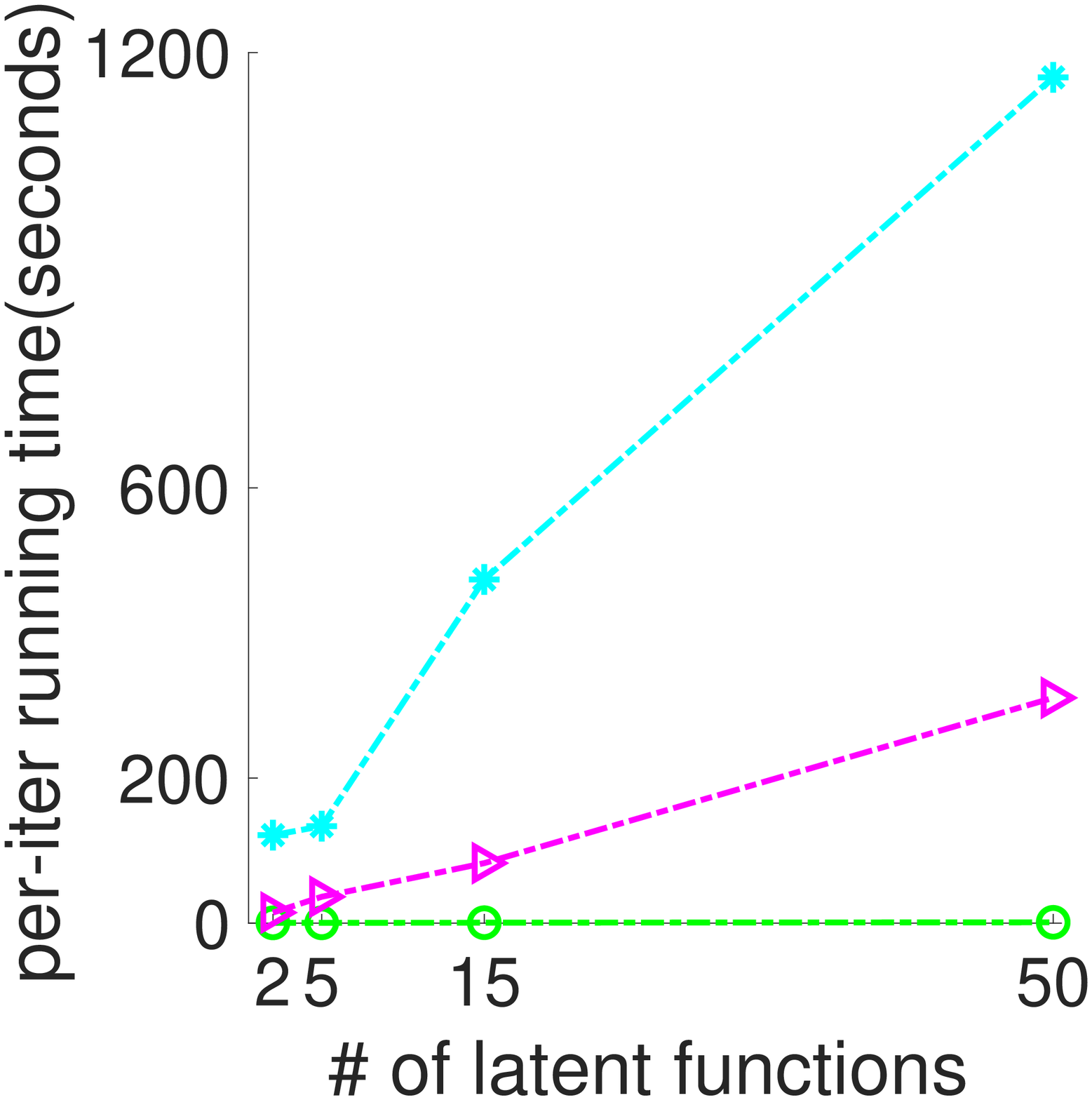}
		\end{minipage}
	}%
	\centering
	\caption{\small Training speed of three GPRN inference algorithms.}
	\label{fig:time_cost}
	\vspace{-0.1in}
\end{figure*}

\subsection{Predicting Small Numbers of Outputs}
We first evaluated the proposed GPRN inference on five real-world datasets with a small number of outputs: (1) \textit{Jura}~\footnote{\url{https://rdrr.io/cran/gstat/man/jura.html}}, the heavy-metal concentration measurements of $349$ neighbouring locations in Swiss Jura. Following \cite{wilson2012gaussian}, we predicted $3$ correlated concentrations, cadmium, nickel and zinc, given the locations of the measurements.  (2)\textit{Equity}~\footnote{https://github.com/davidaknowles/gprn}~\cite{wilson2012gaussian}, a financial datasets that include 643 records of $5$ equity indices --- NASDAQ, FTSE, TSE, NIKKEI and DJC. The inputs are the $5$ indices, and the goal is to predict their $25$ pair-wise correlations. (3) \textit{PM2.5}\footnote{\url{http://www.aqandu.org/}}, 100 spatial measurements (i.e., outputs) of the particulate matter pollution (PM2.5) in Salt Lake City in July 4-7, 2018. The inputs are time points of the measurements. (4) \textit{Cantilever}~\cite{andreassen2011efficient}, material structures with the maximum stiffness on bearing forces from the right side. The input of each example is the force and the outputs are a 3,200 dimensional vector that represents the stress field that determines the optimal material layout in a $80\times 40$ rectangular domain. (5) \textit{GeneExp}\footnote{https://www.synapse.org/\#!Synapse:syn2787209/wiki/70350}, expressions of 4,511 genes (outputs) measured by different microarrays, each of which is described by a $10$ dimensional input vector. 


\paragraph{Competing methods.} We compared our inference algorithm, denoted by SGPRN, with the following approaches. (1) MFVB --- mean-field variational Bayes inference for GPRN ~\cite{wilson2012gaussian}, and (2) NPV --- nonparametric variational Inference for GPRN ~\cite{nguyen2013efficient}. In addition, we compared with three other multi-output GP models that are scalable to high-dimensional outputs: (3) PCA-GP~\cite{higdon2008computer} that uses PCA to find the projection matrix in the LMC framework~\cite{goulard1992linear} for multi-output regression, (4) KPCA-GP~\cite{xing2015reduced} and (5) IsoMap-GP~\cite{xing2016manifold} that use KPCA~\cite{scholkopf1998nonlinear} and (5) IsoMap~\cite{balasubramanian2002isomap} respectively to identify the projection matrix. We also compared with (5) HOGP ~\cite{zhe2019scalable}, high-order GP for regression that introduces latent coordinate features for tensorized output prediction (see Sec. 4 Related Work). 

\begin{figure*}
	\centering
	\subfigure{
		\begin{minipage}[t]{0.3\linewidth}
			\centering
			\includegraphics[width=2in]{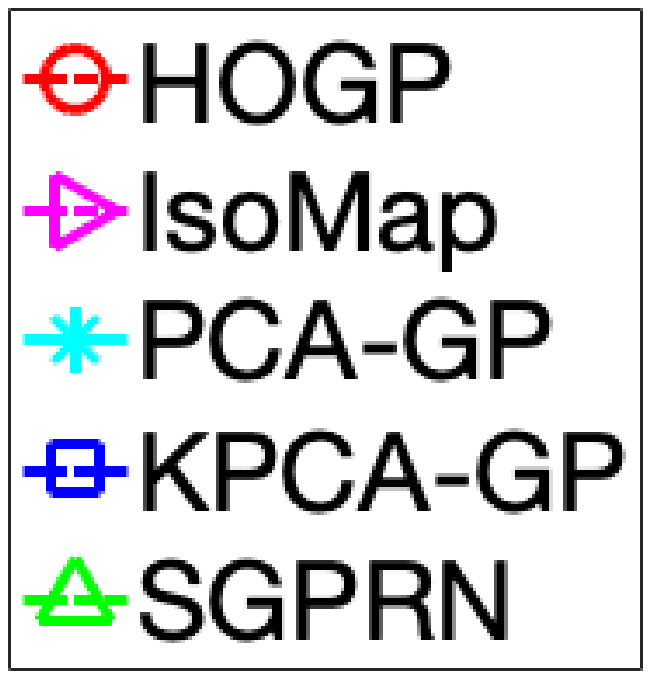}
		\end{minipage}%
	}%
\setcounter{subfigure}{0}
	\subfigure[ \textit{Cantilever} (\# training samples = 128)]{
		\begin{minipage}[t]{0.3\linewidth}
			\centering
			\includegraphics[width=2in]{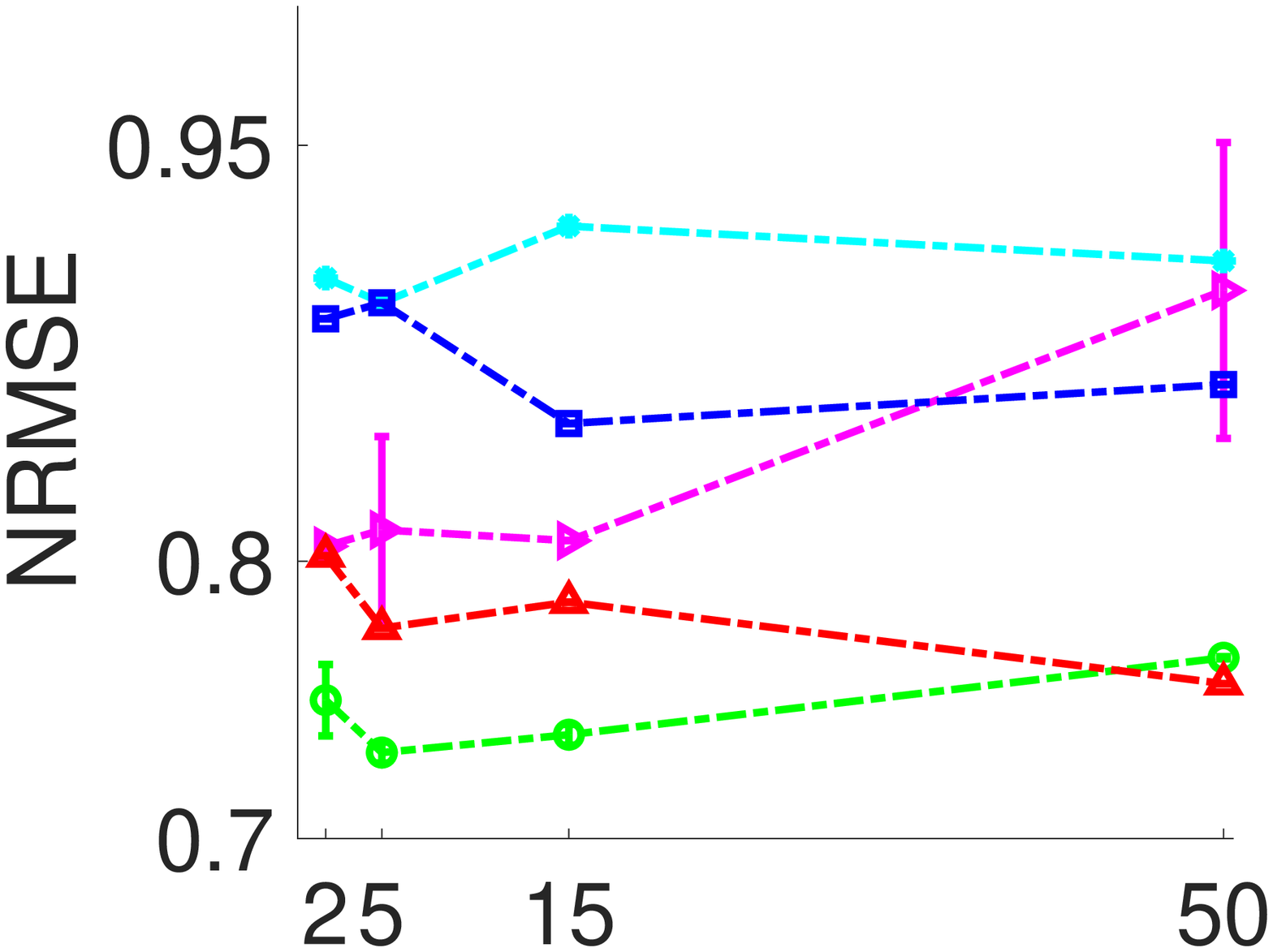}
		\end{minipage}%
	}%
	\subfigure[ \textit{Cantilever} (\# training samples = 256)]{
		\begin{minipage}[t]{0.3\linewidth}
			\centering
			\includegraphics[width=2in]{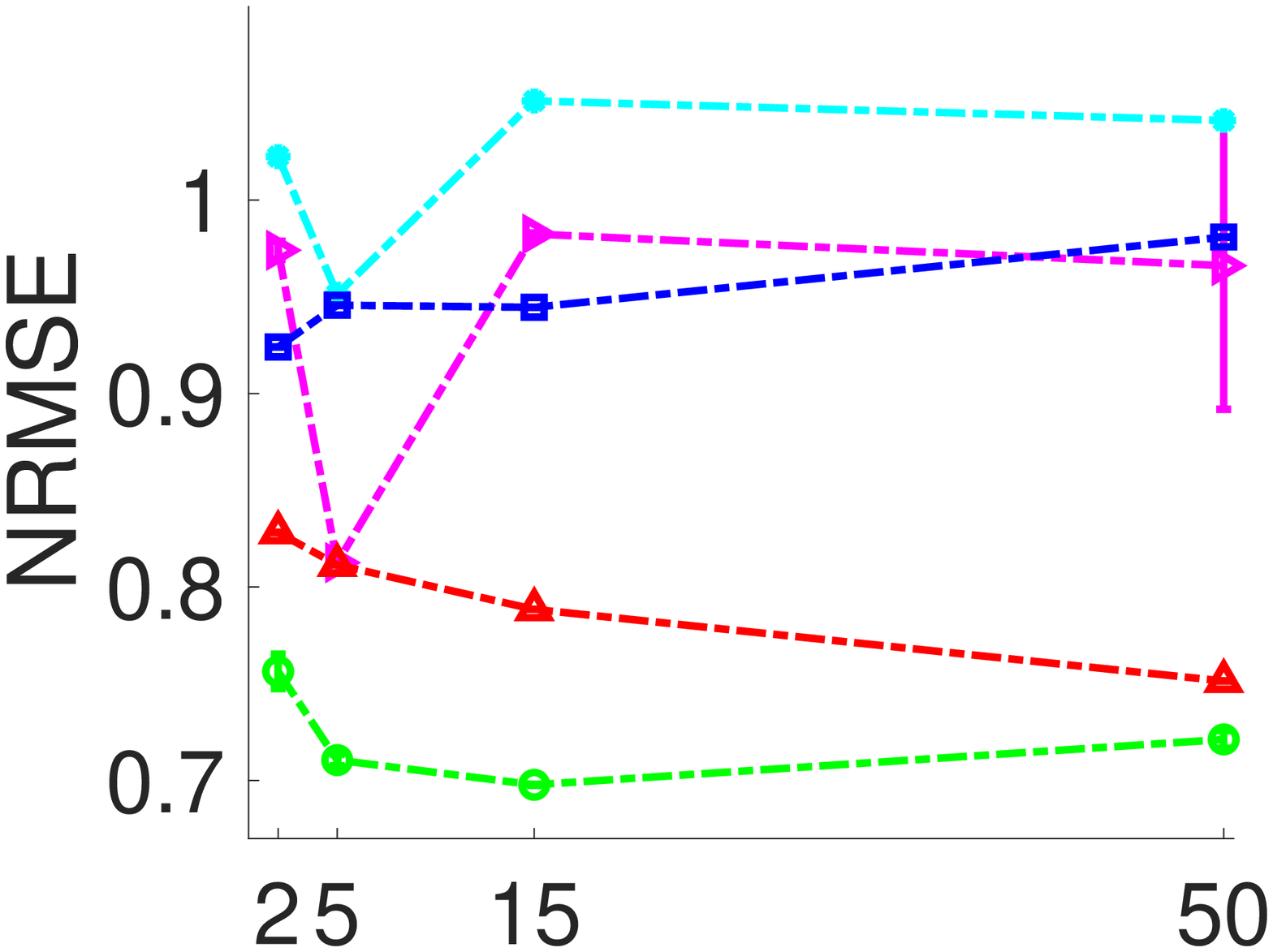}
		\end{minipage}%
	}%
	
	\subfigure[\textit{GenExp} (\# training samples = 128)]{
		\begin{minipage}[t]{0.3\linewidth}
			\centering
			\includegraphics[width=2in]{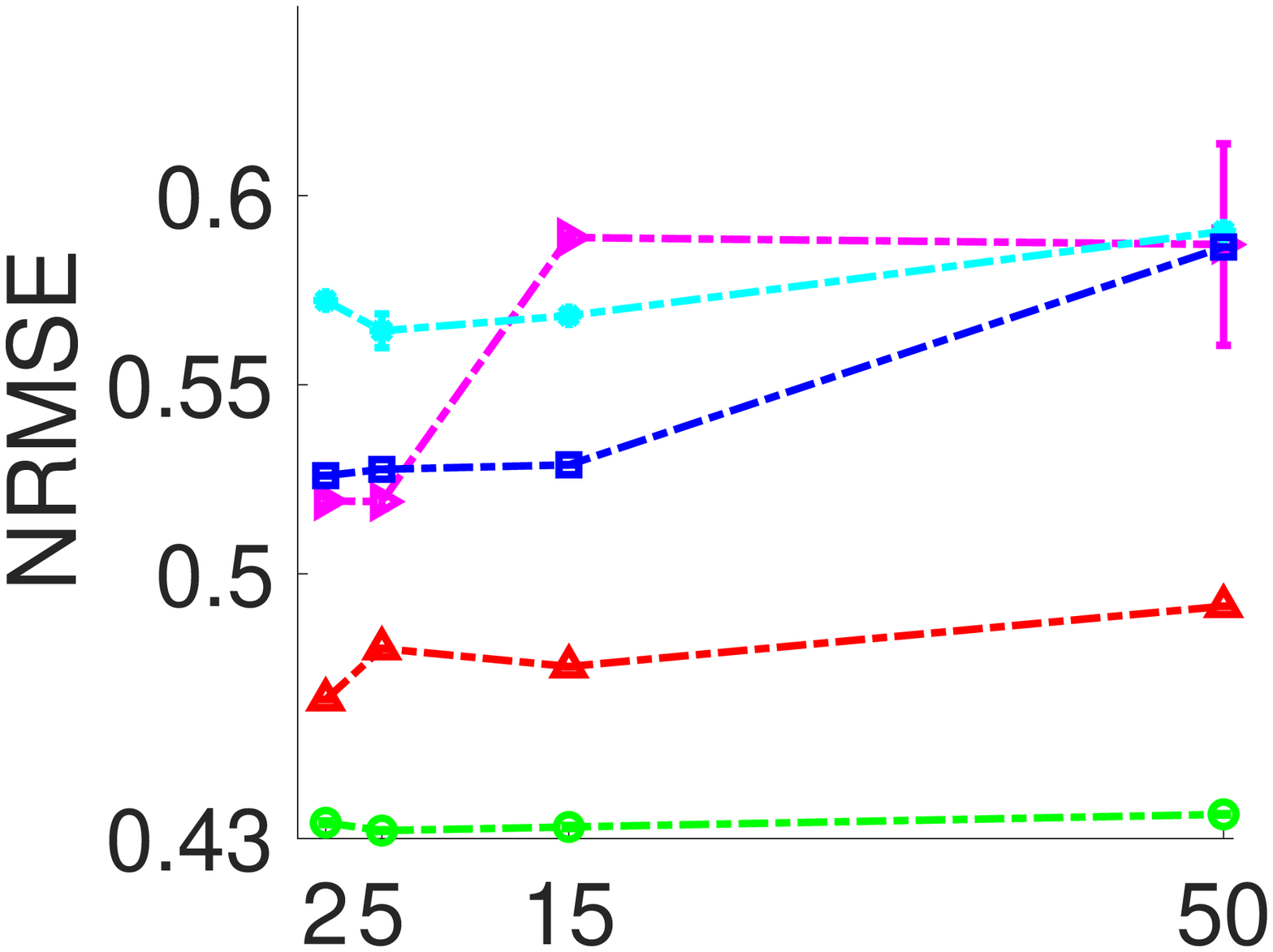}
		\end{minipage}
	}%
	\subfigure[\textit{GenExp} (\# training samples = 256)]{
		\begin{minipage}[t]{0.3\linewidth}
			\centering
			\includegraphics[width=2in]{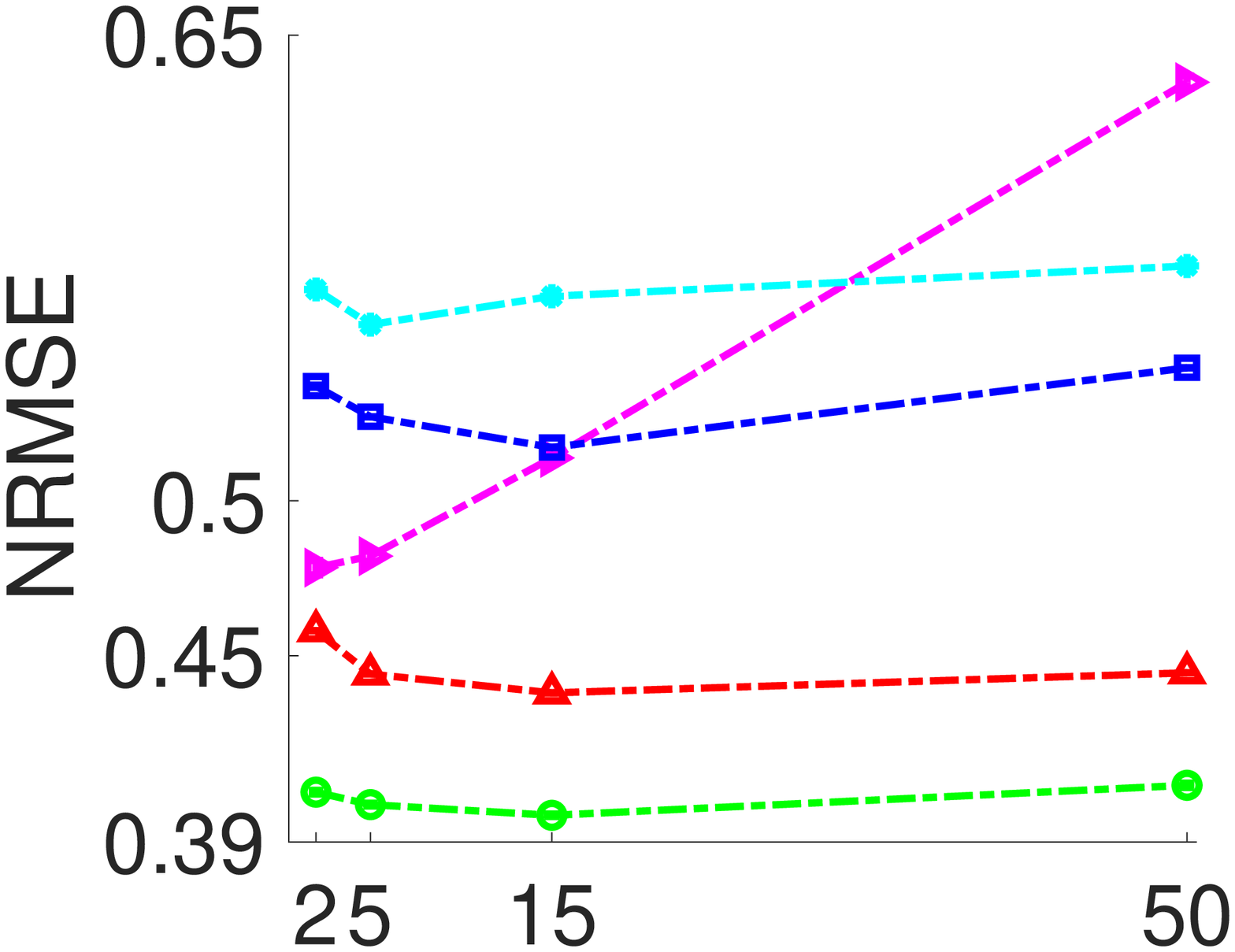}
		\end{minipage}
	}%
	\subfigure[\textit{Pressure} (\# training samples = 64)]{
		\begin{minipage}[t]{0.3\linewidth}
			\centering
			\includegraphics[width=2in]{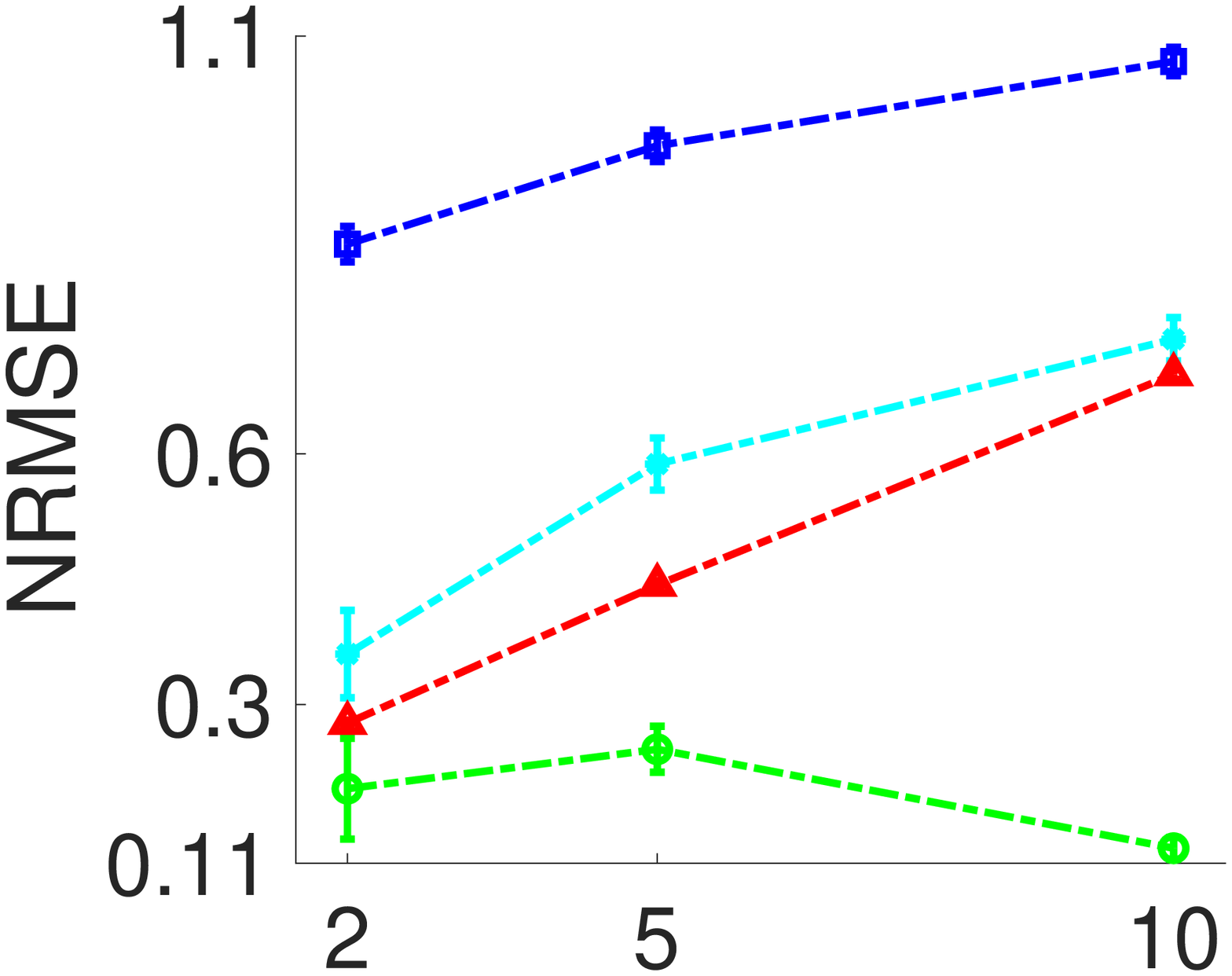}
		\end{minipage}
	}%
	\centering
	\caption{\small The normalized root-mean-squared error(NRMSE) of all the methods on three datasets. The results were averaged over 5 runs. The x-axis represents the number of latent functions of SGPRN, features of HOGP, and bases of IsoMap-GPR, KPCA-GPR, and PCA-GPR.}
	\label{fig:middle_scale}
\end{figure*}

\paragraph{Parameter settings.} We implemented our method SGPRN with TensorFlow~\cite{abadi2016tensorflow}. We used Adam~\cite{kingma2014adam} algorithm for gradient-based optimization and the learning rate was set to $10^{-3}$.\cmt{ and the number of epochs was set to $2,000$.} All the competing methods were implemented with MATLAB. For MFVB and NPV, we used the efficient implementation~(\url{https://github.com/trungngv/gprn}) of the paper~\cite{nguyen2013efficient}. We used $2$ mixture components for the variational poster in NVP (see \eqref{eq:np-post}). The competing methods used L-BFGS for optimization and the number of iterations was set to $100$. We used RBF kernel for all the methods. The input features of all the datasets were normalized, and the kernel parameters (i.e., the length-scale) were initialized to $1$. We varied the number of latent functions/features/bases from $\{2, 5, 15, 50\}$.  
\begin{table}[!htbp]
	\centering
	\begin{tabular}{@{}cc@{}}
		\toprule
		\textit{Jura}     &             \\ \midrule
		SGPRN & \textbf{0.5127}$\pm$0.002   \\
		MFVB  & 0.5237$\pm$0.001            \\
		NPV   & 0.6088$\pm$9e-06            \\ \midrule
		\textit{Equity} &               \\ \midrule
		SGPRN & \textbf{2.5759e-05}$\pm$2e-07 \\
		MFVB  & 4.4530e-05$\pm$6e-07          \\
		NPV   & 4.7267e-05$\pm$9e-18          \\ \midrule
		\textit{PM2.5}     &           \\ \midrule
		SGPRN & \textbf{1.07089}$\pm$0.02   \\
		MFVB  & 1.3916$\pm$0.06            \\
		NPV   & 14.7101$\pm$0.14            \\ \bottomrule
	\end{tabular}
	\caption{\small The mean absolute error(MAE) of the three GPRN inference methods. The results were averaged over $5$ runs.}
	\label{table:small_scale}
	\vspace{-0.1in}
\end{table}
\paragraph{Comparison with state-of-the-art GPRN inference.} We first compared with MFVB and NPV on the three smallest datasets, \textit{Jura}, \textit{Equity} and \textit{PM2.5}, with $2$, $25$ and $100$ outputs respectively. We tested SGRPN, MFVB and NPV on a workstation with 2 {Intel(R) Xeon(R) E5-2697} CPUs, 28 cores and 196GB memory. 
Note that MFVB and NPV are not available on the other datasets (i.e., \textit{Cantilever} and \textit{GeneExp}) --- our test shows that they will take extremely long time (weeks or months) to train with $15$ and $50$ latent functions.  We followed the setting of the original paper~\cite{wilson2012gaussian} and ~\cite{nguyen2013efficient} to only use $2$ latent functions. On \textit{Jura}, we randomly split the data into $249$ examples for training and $100$ for test, on \textit{Equity} $200$ for training and $200$ for test, and on \textit{PM2.5} $256$ for training and $32$ for test. In our algorithm, for \textit{Jura} and \textit{Equity}, we used the original output space, and for \textit{PM2.5} we tensorized the output space to be $10 \times 10$. We ran our algorithm for $2,000$ epochs to ensure convergence; both MFVB and NPV converged after $100$ iterations. We repeated for $5$ times and reported the average of the mean absolute error (MAE) and the standard deviation in Table \ref{table:small_scale}. As we can see, our method SGPRN significantly outperforms (p-value $< 0.05$) both MFVB and NPV on all the three datasets, showing superior inference quality. Note that NPV obtains much bigger MAEs on \textit{PM2.5}. This might be because the optimization of the variational ELBO converged to inferior local maximums. 

Next, we compared with MFVB and NPV in terms of the speed. We reported the per-iteration running time of MFVB and NPV in Fig. \ref{fig:time_cost}. Since our algorithm ran $2,000$ epochs while MFVB and NPV $100$ iterations, we used the average running time of $20$ epochs of SGPRN as the per-iteration time for a fair comparison. As we can see, the speed of SGPRN is close to that of MFVB and NPV for very a few latent function ($2$ and $5$). However, with more latent functions, SGPRN becomes much faster. For example, on \textit{PM2.5} with $50$ latent functions, SGPRN gains $200$x  and $785$x speed-up as compared with MFVB and NPV. This is consistent with their time complexities, \{MFVB: $\Ocal(NK^2D)$, NPV: $\Ocal(QN^2KD)$, SGPRN:$\Ocal(NKD)$\}.

\paragraph{Comparison with other methods.} Next, we compared with other scalable multi-output GP regression models. To this end, we used  \textit{Cantilever} and \textit{GeneExp} datasets. For our algorithm and HOGP, we tensorized the output space of \textit{Cantilever} to $80 \times 40$ and \textit{GeneExp} to $347 \times 13$. On each dataset, we randomly chose $\{128, 256\}$ examples for training and $100$ from the remaining set for test. We repeated this procedure for $5$ times and reported the average normalized root-mean-square error (NRMSE) and the standard deviation of each method in Fig. \ref{fig:middle_scale}. As we can see, SGPRN significantly outperforms the competing methods (p-value $<0.05$) in all the cases except when training on \textit{Cantilever} with $128$ examples and $50$ latent functions , SGPRN was a little worse than PCA-PG (Fig. 2a). Therefore, it demonstrates the advantage of GPRN in predictive performance, which might be due to its capability of capturing non-stationary output dependencies.

\subsection{Large-Scale Physical Simulations for Lid-Driven Cavity Flows}
Finally, we applied SGPRN in a large-scale physical simulation application. 
Specifically, we trained GPRN to predict a one-million dimensional pressure field for lid-driven cavity flows~\cite{bozeman1973numerical}, which include turbulent flow inside the cavity. The simulation of the field is done by solving the Navier-Stoke equations that are known to have complicated behaviours under large Reynolds numbers. We used a fine-grained mesh to ensure the numerical solver to converge. The input of each simulation example is a $5$ dimensional vector that represents a specific boundary condition. The computation for each simulation is very expensive, so we only collected $96$ examples. Hence, this is a typical ``large $D$, small $N$'' problem. We randomly split the dataset into $64$ training and $32$ test examples, and then ran SGPRN, PCA-GP, KPCA-GP and IsoMAP-GP. For SGPRN, we tensorized the one million outputs into a $100 \times 100 \times 100$ tensor. We varied the number of latent functions from $\{2, 5, 10\}$. We repeated the training and test procedure for $5$ times and showed the average normalized root-mean-square error (NRMSE) of each method in Fig.\ref{fig:middle_scale}(e). As we can see, our method significantly outperforms all the competing approaches,  by a large margin especially when using $5$ and $10$ latent functions. The results further demonstrate the advantage of GPRN for large-scale multi-out regression tasks when it is available.

\vspace{-0.05in}
 \section{Conclusion}
We have proposed a scalable variational inference algorithm for GPRN, a powerful Bayesian multi-output regression model. Our algorithm not only improves the inference quality upon the existing GPRN inference methods but also is much more efficient and scalable to a large number of outputs. In the future work, we will continue to explore GPRN in large-scale multi-output regression applications.   

\section*{Acknowledgment}
This work has been supported by DARPA TRADES Award HR0011-17-2-0016 and NSF IIS-1910983.

\bibliographystyle{named}
\bibliography{SGPRN}

\end{document}